\documentclass[10pt,twocolumn,letterpaper]{article}

\usepackage{cvpr}
\usepackage{times}
\usepackage{epsfig}
\usepackage{graphicx}
\usepackage{amsmath}
\usepackage{amssymb}
\usepackage{pifont}

\usepackage[pagebackref=true,breaklinks=true,letterpaper=true,colorlinks,bookmarks=false]{hyperref}

\cvprfinalcopy 

\def\cvprPaperID{213} 

\ifcvprfinal\fi
\begin{document}

\title{Flash Photography for Data-Driven Hidden Scene Recovery}

\author{Matthew Tancik, Guy Satat, Ramesh Raskar\\
MIT Media Lab\\
75 Amherst st. Cambridge MA, USA\\
{\tt\small \{tancik, guysatat\}@mit.edu, ramesh@media.mit.edu}
}

\maketitle

\begin{abstract}
   Vehicles, search and rescue personnel, and endoscopes use flash lights to locate, identify, and view objects in their surroundings. Here we show the first steps of how all these tasks can be done around corners with consumer cameras. Recent techniques for NLOS imaging using consumer cameras have not been able to both localize and identify the hidden object. We introduce a method that couples traditional geometric understanding and data-driven techniques. To avoid the limitation of large dataset gathering, we train the data-driven models on rendered samples to computationally recover the hidden scene on real data. The method has three independent operating modes: 1) a regression output to localize a hidden object in 2D, 2) an identification output to identify the object type or pose, and 3) a generative network to reconstruct the hidden scene from a new viewpoint. The method is able to localize 12cm wide hidden objects in 2D with 1.7cm accuracy. The method also identifies the hidden object class with 87.7\% accuracy (compared to 33.3\% random accuracy). This paper also provides an analysis on the distribution of information that encodes the occluded object in the accessible scene. We show that, unlike previously thought, the area that extends beyond the corner is essential for accurate object localization and identification. 
\end{abstract}


\section{Introduction}

\begin{figure}[t]
\includegraphics[width=\columnwidth]{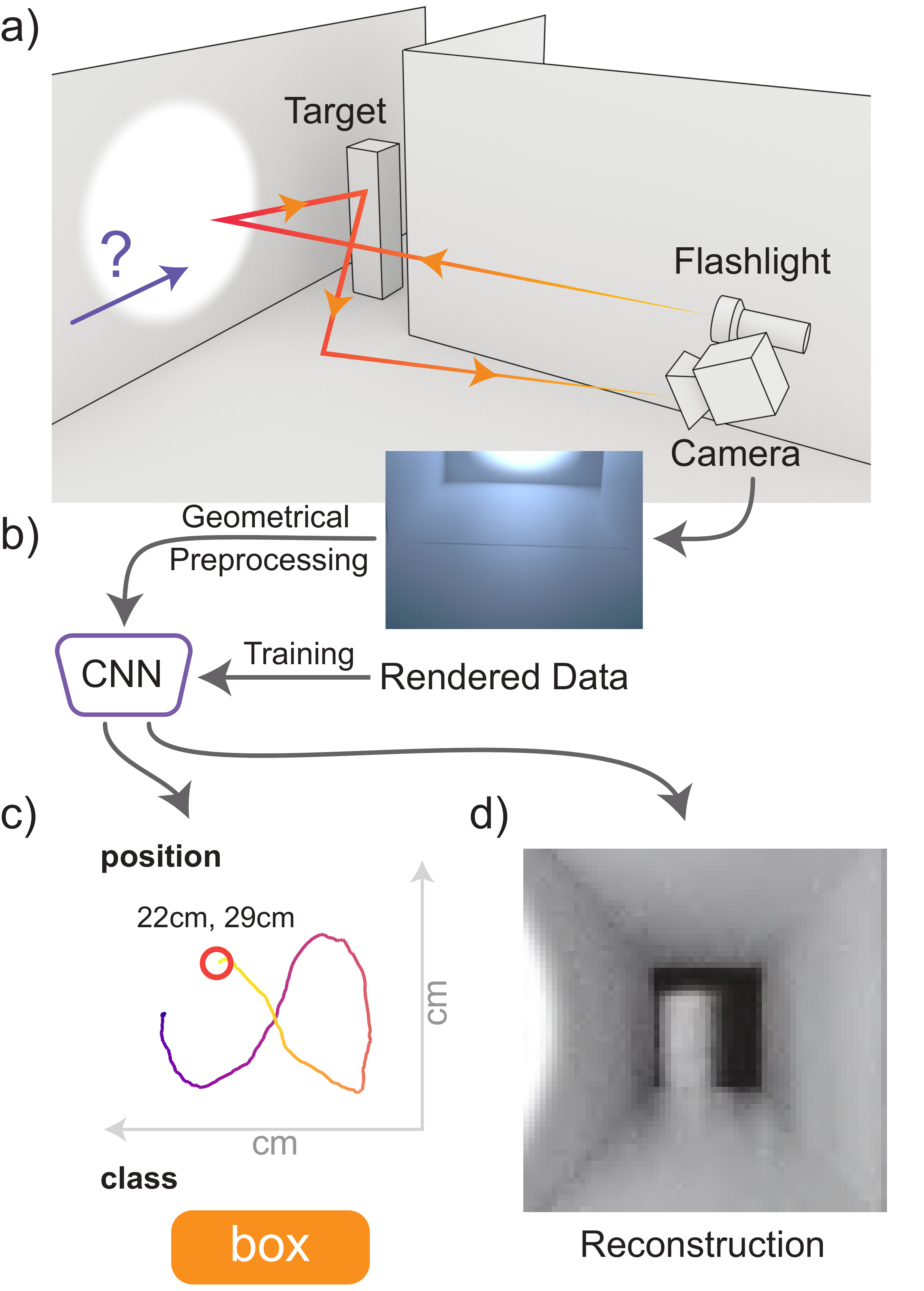}
\centering
\caption{Can we create an image as if seen from the '?' using self contained consumer flash photography? a) A target is occluded from the camera. The scene is illuminated by a flashlight, light scatters off of the wall to the occluded object, floor, and to the camera. The measured signal changes depending on the class and location of the hidden target. b) The Measurement is processed and fed into a CNN for inference. The CNN is trained using rendered data. c) The system localizes, identifies and tracks the occluded object in real time. d) A generative model reconstructs a photo-realistic rendering of the hidden scene from the desired virtual camera's point of view.}
\label{fig:overview}
\end{figure}

\begin{table*}[t]
\centering
\caption{Comparison of NLOS imaging techniques with a consumer camera.}
\label{tab:comparison}
\includegraphics[width=\textwidth]{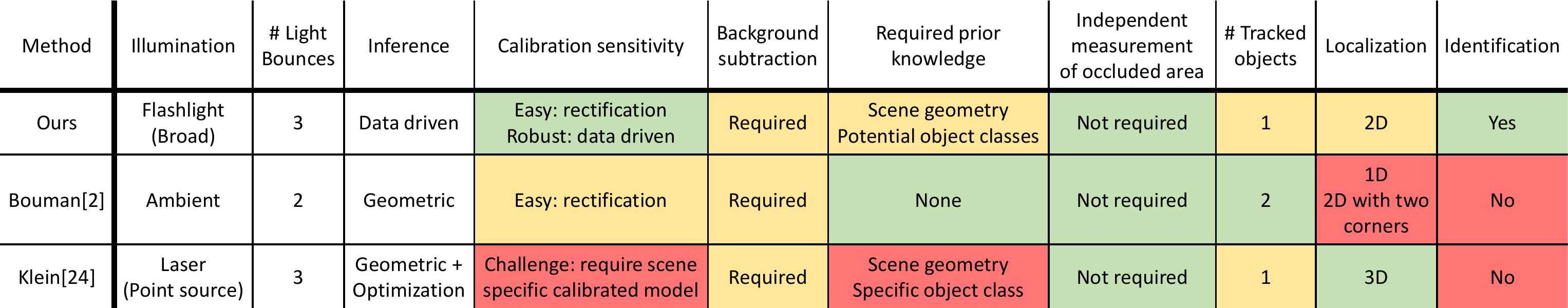}
\end{table*}

The use of active illumination is common in computer and human vision. A car is equipped with headlights to help the driver or autonomous controller to drive at different times of day. A firefighter uses a flashlight or headlamp to see his surroundings while walking in a dark corridor. An endoscope uses bright illumination to navigate inside the body. Our goal in this paper is to extend these capabilities beyond the line of sight using consumer flash photography and data-driven recovery algorithms.  

The task of seeing around corners has been accomplished with various measurement devices~\cite{Velten2012, Heide2014c, Gariepy2015a, Kadambi2016}. Recently it was demonstrated using a traditional consumer camera with~\cite{Klein2016} and without~\cite{Bouman_2017_ICCV} an active illumination source. Various computer vision applications have been demonstrated for non-line-of-sight (NLOS) imaging such as full scene reconstruction~\cite{Velten2012}, tracking~\cite{Gariepy2015a}, classification~\cite{satat2017object}, and localization~\cite{caramazza2017neural}.

To improve robustness of NLOS imaging with consumer flash photography, and demonstrate new capabilities like hidden object identification and photo-realistic hidden scene reconstruction, we turn to data-driven methods which have already proved useful in NLOS imaging~\cite{satat2017object, caramazza2017neural}. As in other computer vision applications, the main advantage of using such techniques is the inherent robustness and the fact that an accurate physical model is not necessary. Furthermore, this data-driven approach is a step toward general purpose hidden object classification.

The main challenge for data-driven solutions in the domain of NLOS imaging is the lack of large labeled databases. To tackle this challenge, we render a dataset for NLOS imaging, which is only limited by computational resources. To overcome the limitation of transfer learning (ability to train on rendered data and operate on real-world data) we help the network to generalize by: 1) combining classical geometric knowledge into our imaging pipeline, and 2) introducing variations in rendering parameters into our dataset. 

Here, we experimentally demonstrate NLOS imaging in a hallway with a consumer camera, coupled with a flashlight and a data-driven recovery algorithm. The main contributions of this paper are:
\begin{enumerate}
\item A technique to identify, localize, and track in 2D a hidden object around a corner using a combination of classical geometric processing and a data-driven approach.
\item A technique for full scene reconstruction that recovers the scene from a point of view that is not accessible to the camera based on a generative model.
\item An analysis of the spatial distribution of information on the occluded object in the visible scene.
\end{enumerate}

\section{Related Works}

\subsection{Imaging Beyond Line of Sight}
NLOS imaging has been explored widely in the context of vision related applications such as: image dehazing~\cite{Berman,Cai2016,Fattal2008,Schechner2003}, seeing through translucent objects~\cite{Fuchs2008}, cloud tomography~\cite{Holodovsky2016,Levis2015}, underwater imaging~\cite{Sheinin2016}, and tracking~\cite{Gariepy2015a,Iiyama2014}. It has also been discussed in the context of structured light~\cite{Narasimhan2005}. 

Seeing around a corner requires solving an elliptical tomography problem which involves both the geometry and albedo. This challenge has been met by restriction of the problem,  and solution with, e.g., backpropagation~\cite{Velten2012} or sparsity priors~\cite{Kadambi2016}. A different approach uses only ballistic photons to recover geometry~\cite{tsai2017geometry}.  

Different geometries of NLOS imaging have been demonstrated including around corners~\cite{Velten2012}, and through thin and volumetric scattering media~\cite{Satat2016a}. Finally, various applications have been considered like full scene reconstruction~\cite{Jin2015a}, pose estimation~\cite{satat2017object}, identification~\cite{caramazza2017neural}, tracking~\cite{Gariepy2015a}, and reconstruction of the hidden room shape~\cite{pediredla2017reconstructing}.

NLOS imaging has been explored with different hardware. Most techniques are based on active illumination and time-of-flight sensors. These can be either pulsed~\cite{Buttafava2015a, Laurenzis2015b, Velten2012} or AMCW~\cite{Heide2014c,Heide2014b,Kadambi2016} based. 

Recently, regular cameras have been used for NLOS imaging. A method by Klein et al.~\cite{Klein2016} uses a laser pointer and a regular camera to track objects around a corner. A passive method (relies on ambient light) by Bouman et al.~\cite{Bouman_2017_ICCV} is able to perform one dimensional tracking of objects around a corner. Table~\ref{tab:comparison} summarizes the key differences between our approach and these two techniques. Two key properties of our approach are advantageous: 1) using flash photography is appealing for various practical scenarios as described before, and 2) our reliance on a data-driven technique allows us to demonstrate multiple computer vision tasks. Specifically, we locate the target in 2D, identify the object, and reconstruct a photo-realistic image of the hidden target area. Furthermore, since our approach is data-driven, it does not require extensive calibration similar to the one required by Klein et al. The localization accuracy reported by Klein et al. is 6.1cm (in 3D) compared to our reported accuracy of 1.7cm (in 2D). Bouman et al. provided error ellipses on their plots without explicitly specifying localization accuracies.

\begin{figure}[t]
\includegraphics[width=0.8\linewidth]{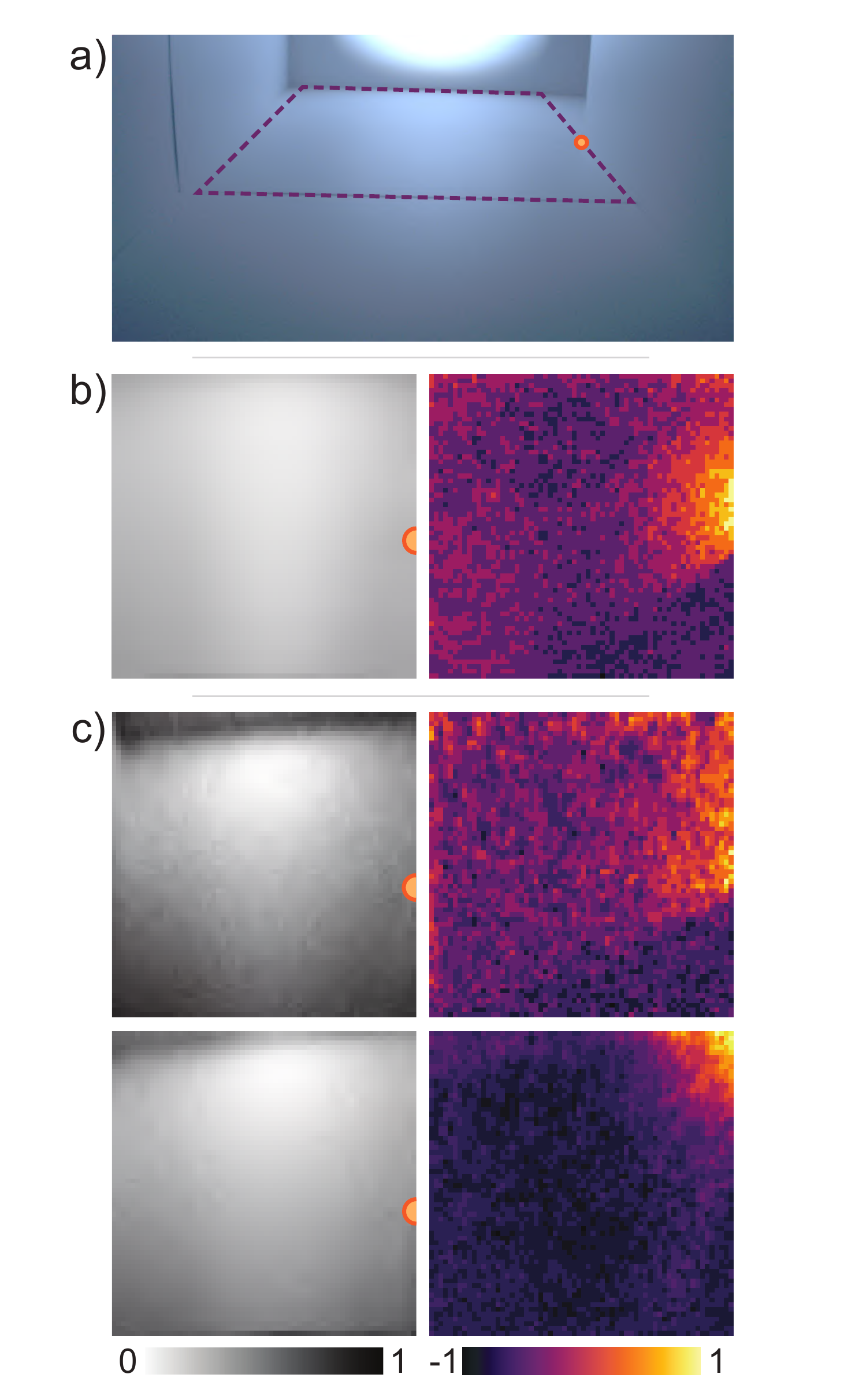}
\centering
\caption{Geometric preprocessing. a) Photograph from camera before processing. Purple dotted line outlines the floor area to be rectified. The orange circle indicates the location where the corner wall meets the floor. b) The photograph is first rectified and cropped (left column, grayscale). Then, it is background subtracted and scaled to $[-1,1]$ (right column, color). This is the input to the CNN. c) Two examples taken from the rendered dataset. The background subtracted examples demonstrate the light casted by the object which spreads throughout the measured area, not just the edge discontinuity.}
\label{fig:datasets}
\end{figure}

\subsection{Data-Driven Approaches to Imaging}
Data-driven techniques have also been applied to problems in the computational imaging community such as phase imaging~\cite{sinha2017lensless}, compressive imaging~\cite{kulkarni2016reconnet}, tomography~\cite{kamilov2015learning}, microscopy~\cite{horstmeyer2017convolutional}, imaging~\cite{horisaki2016learning}, and imaging through scattering media~\cite{satat2017object}. Localization and identification of people around a corner has also been demonstrated with a data-driven approach based on active pulsed illumination and a time-resolved single photon avalanche diode camera~\cite{caramazza2017neural}.

Using rendered data to train CNN models is becoming more common in computer vision applications such as optical flow~\cite{Sintel}, action recognition~\cite{PHAV}, overcoming scattering~\cite{satat2017object}, text recognition~\cite{SynthText}, and tracking~\cite{VirtualKITTI}.

Here, we demonstrate the use of CNNs for NLOS imaging with a consumer camera. We show that rendering data using a graphics engine can be used to train a model that performs inference on real data. Finally, we demonstrate that a generative model can be used to reconstruct an occluded scene from a new point of view.

\section{Localizing and Identifying Objects Beyond Line of Sight}

As objects are added or removed from a scene, the measured irradiance changes with areas becoming dimmer or brighter. These changes still occur even when the object is moving in an occluded part of the scene. However, they are usually too subtle for the human eye to perceive, as our eyes are constantly adapting to our surroundings and ignore such information. In some cases, the changes in brightness are perceptible, but it can be hard to discern the cause of such changes. In either case this phenomenon is generally amplified with controlled illumination like flash photography. This section describes a technique to leverage such changes to localize and identify the hidden object.

In our experiments we are solving the problem of an 'L' shaped hallway. The corner is occluding an object from the camera as shown in Fig.~\ref{fig:overview}a. A flashlight is adjacent to the camera and is pointing at the wall opposite the camera. The camera is focused on the floor near the corner. In this arrangement, the useful signal is when light follows the path: light source $\rightarrow$ wall $\rightarrow$ object $\rightarrow$ floor $\rightarrow$ camera. Fig.~\ref{fig:overview}a shows the geometry and example light path in the scene.

\begin{figure*}[t]
\includegraphics[width=\textwidth]{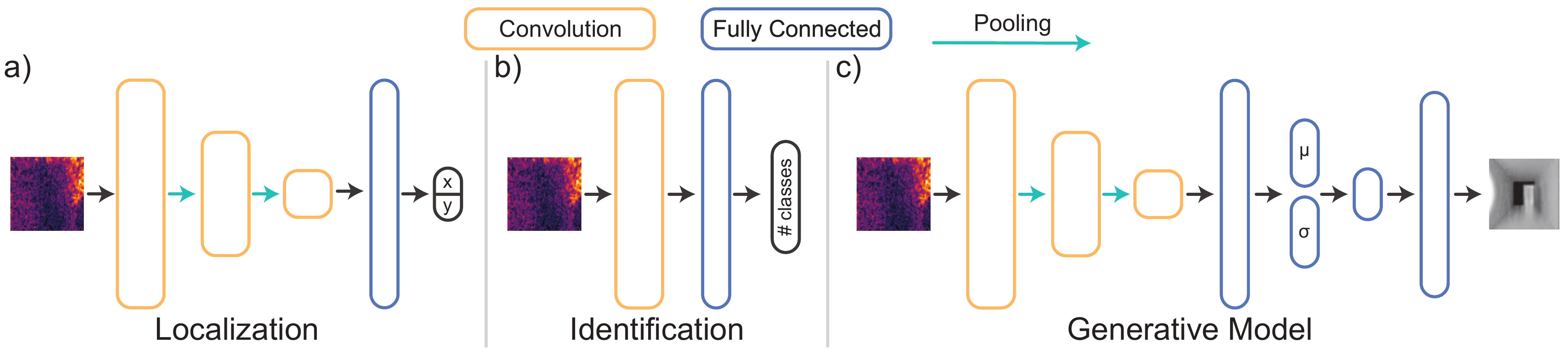}
\caption{Network architectures for localization, identification, and generative model. The input to all networks is $64 \times 64$ pixels scaled to $[-1,1]$. Convolution layers (yellow squares) are of size $5 \times 5$ with a stride of 1. Pooling layers (green arrows) are of size $2 \times 2$. A RELU is performed after each pooling layer and before each fully connected layer (blue square). a) The localization network has two outputs. b)~ The identification network uses softmax for classification. c) The generative model has a VAE architecture where the decoder is a series of fully connected layers. The embedding vector is composed of the mean (length 100) and variance (length 100) of a probability distribution, the latent vector is of length 20.}
\label{fig:cnns}
\end{figure*}

\subsection{Rendering Synthetic Dataset}
\label{sec:sim_datasets}
Many of the successes of deep learning systems in computer vision can be attributed to the availability of large scale datasets. The creation of such datasets is often difficult, time consuming, or impractical. For these reasons, our method utilizes rendered data for training and real data for testing and analysis. There are several advantages in using rendered data: First, the size of the dataset is bounded only by computation. Second, it is easy to increase diversity in the dataset by varying scene parameters such as material properties. And third, the labels are used in the generation of the data and are thus exact (noiseless labels).

To render photo-realistic measurements that account for the interaction of light with the hidden object we use a physics-based ray tracing engine (Blender Cycles). To improve system robustness, the diversity of the dataset is increased by varying rendering parameters such as illumination, camera properties, material properties, and scene geometry (further data generation details are provided in the supplement). 

The training dataset consists of $75,000$ monochromatic images of $64 \times 64$ pixels. Generating a single image takes $\sim 5$ seconds on an Nvidia GTX 1080 GPU. Examples of the rendered samples are shown in Fig.~\ref{fig:datasets}.

\subsection{Geometric preprocessing}
\label{sec:geom_features}

\textbf{Rectification}\qquad 
To increase robustness and allow the algorithm to operate independently of the camera position we first rectify the input image to a top-down view of the floor. The rectification involves placing the corner at the same pixel, thus easing the network to learn a model that is invariant to the camera's point of view. Since this rectification process is rather simple, it is performed using traditional computer vision geometry. To rectify, we select four points, two on the back wall and two a known distance down the hallway closer to the camera. This quadrilateral is shown in Fig.~\ref{fig:datasets}a. The homography is calculated and used to project the quadrilateral into a $64 \times 64$ pixel square. Another approach would be to learn such transformation~\cite{detone2017toward}.

\textbf{Background Subtraction}\qquad To further amplify the signal in the images we subtract the background from the measurements. Background subtraction has been a requirement in previous NLOS imaging when using consumer cameras~\cite{Bouman_2017_ICCV,Klein2016}. We explore several background subtraction techniques, including ground truth subtraction~\cite{Klein2016}, subtraction of pixel-wise mean~\cite{Bouman_2017_ICCV}, and pixel-wise minimum of an input video. We note that all approaches require knowledge about the hidden scene, for ground truth we need to know the object is not there, and for mean or minimum subtraction we require the object to substantially move in the hidden scene during the measurement time. In our experiments, we found that subtraction of the ground truth provides better accuracies: ground truth subtraction: 1.71cm, 87.7\%, mean subtraction: 12.26cm, 38.9\%, minimum subtraction: 5.37cm, 79.2\% for localization and identification accuracies respectively. Therefore results provided throughout this paper are based on ground truth background subtraction. Fig.~\ref{fig:datasets} demonstrates raw measurement and synthetic rendering after rectification and background subtraction.

\subsection{Localization and Identification Models}
\label{sec:local_model}

Two CNN models were trained for object localization and identification. The input image (following rectification and background subtraction) was scaled to $[-1,1]$. Scaling individual images was necessary as the standard procedure of subtracting the dataset mean did not work in our case, due to the differences of the statistics between the real and simulated data.

\textbf{Localization}\qquad  The localization model is a CNN regression network trained to predict the $(x,y)$ location of the object. The architecture of the model is diagrammed in Fig.~\ref{fig:cnns}a. The model was trained using a mean square error loss function for 7 epochs. 

\textbf{Identification}\qquad A second model of similar architecture was trained to identify the type of the object. A softmax layer was used for identification, and training was based on the cross entropy loss for 20 epochs.

\begin{figure}[t]
\includegraphics[width=.95\linewidth]{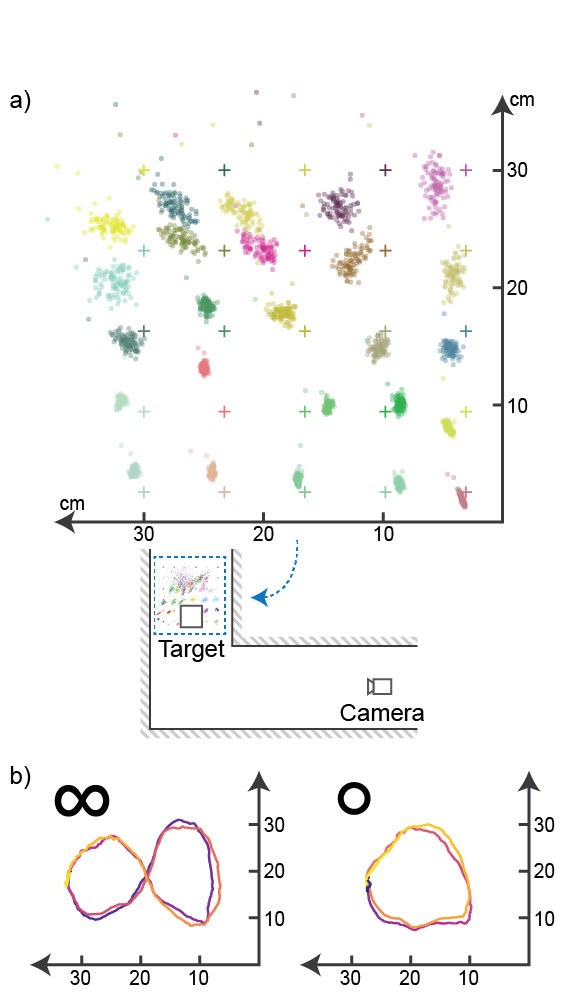}
\centering
\caption{Occluded object localization and tracking in 2D. a) Localization accuracy on a grid. Three objects of different shapes are moved along a $5 \times 5$ grid of size $30 \times 30$ $cm^2$. The center of each grid point is marked by a colored plus sign. Each object and each grid point are measured 30 times, during which the object is perturbed (90 data points per grid point, 2250 data points overall). Localizations are marked by colored dots. The color of each dot corresponds to the associated ground truth grid position, such that correct localizations have the same color as the closest ground truth plus sign. As the object is further away, the localization accuracy is reduced. The inset shows the hallway geometry, with the localizations superimposed and an example of one of the targets for scale reference. b)  Tracking a hidden object. In these examples an object is tracing an infinity sign and circular path twice. Each frame in the camera video stream is processed and used for localization independently. Color along the path represents the time axis.}
\label{fig:local}
\end{figure}

\subsection{Experimental Results}
\label{sec:exp_res}
To evaluate the method, an 'L' shaped hallway scene was constructed using poster boards. A set of objects of different shapes were placed in the occluded area. A Flea3 Point Grey camera captured 16bit monochromatic images at 30fps. The scene was illuminated by an LED flashlight. Examples of the real world samples can be seen in Fig.~\ref{fig:datasets}b. The supplement provides further experimental details.

A test dataset was gathered in the above settings to evaluate the localization and identification accuracy. Three different objects (rectangular box with a base of 12cm, cube with a base of 18cm, and pyramid with a base of 12cm) were placed in 25 different positions on a uniform $30 \times 30 cm^2$ grid. The grid starts 10cm away from the corner. At each position, and for each object, 30 images were captured and processed (90 samples per grid point, totaling 2250 examples in the test dataset). In addition, the objects were jittered during acquisition.

\subsubsection{Localization and Tracking}
Fig.~\ref{fig:local}a shows the computed locations alongside the ground truth. As can seen from the localization distributions, when the object is close to the visible area the localization accuracy is higher (1.71 cm accuracy when the object is within 25 cm from the corner). When the object retreats further away the accuracy reduces (3.21 cm accuracy when the object is between 25-45 cm from the corner). The overall localization accuracy is 2.61cm across all examples in the test set. Another notable localization limitation is when the object is closer to the wall being illuminated. In that case, it is less likely to be properly illuminated by the reflection from the wall itself and thus contributes less to the measured signal on the floor. This in turn results in reduced localization accuracy.

Since the required computational steps are simple geometrical operations and a single inference pass through a shallow network, the method operates in real time. We track the object over time by localizing it in each frame independently. To that end, we recorded a video of the object tracing in an infinity sign and a circular motion for a total of 500 frames. The predicted locations are presented in Fig.~\ref{fig:local}b (note that these tracking plots did not require any smoothing or filtering). A supplemental video demonstrates real time tracking.

\begin{figure}
\includegraphics[width=\linewidth]{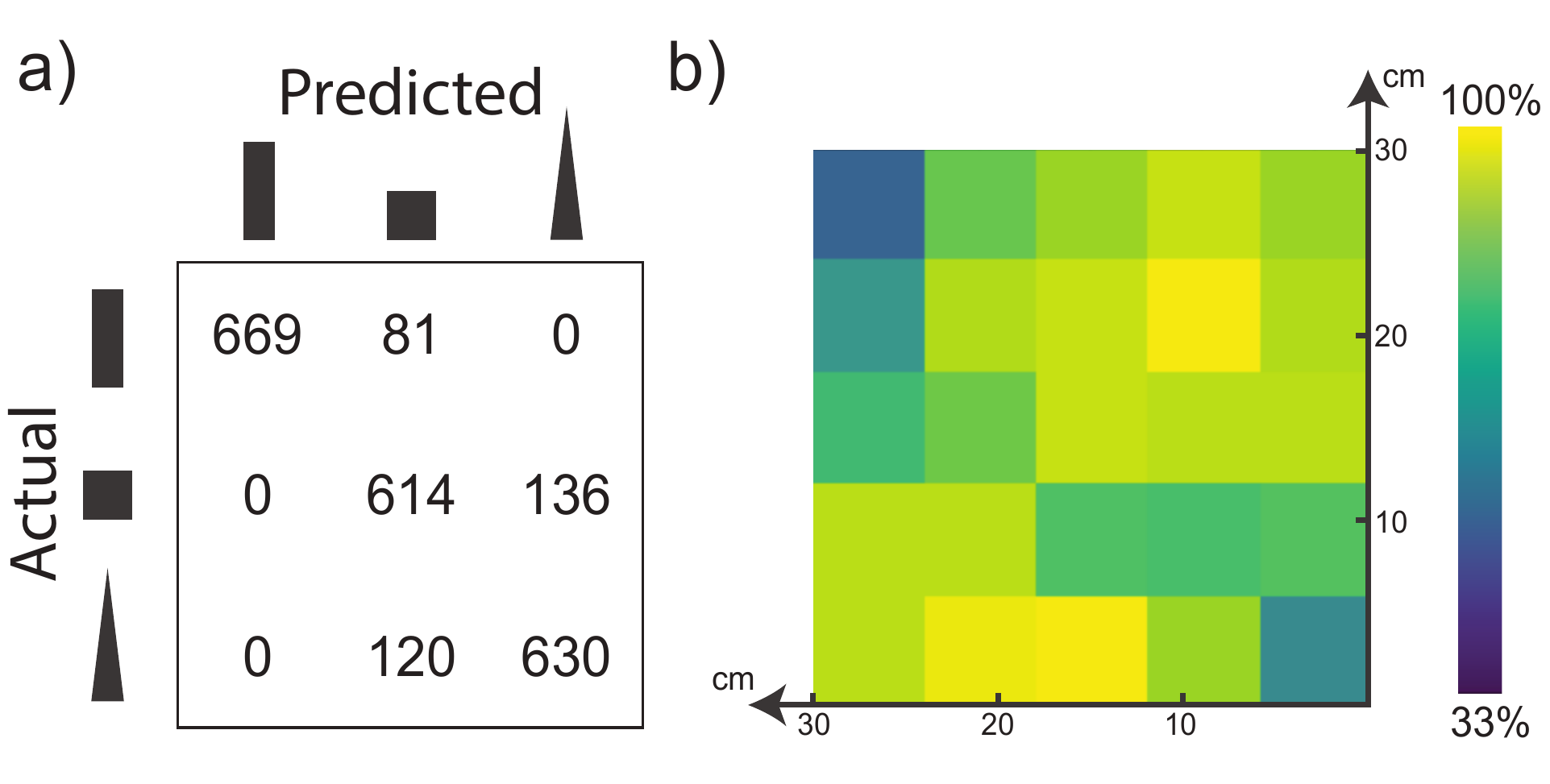}
\caption{Hidden Object identification. a) A confusion matrix showing the identification accuracies among the three classes on real data. The reported accuracies are averages over all locations. b) Identification accuracies on different grid positions. }
\label{fig:class}
\end{figure}

\begin{figure*}[t]
\includegraphics[width=\textwidth]{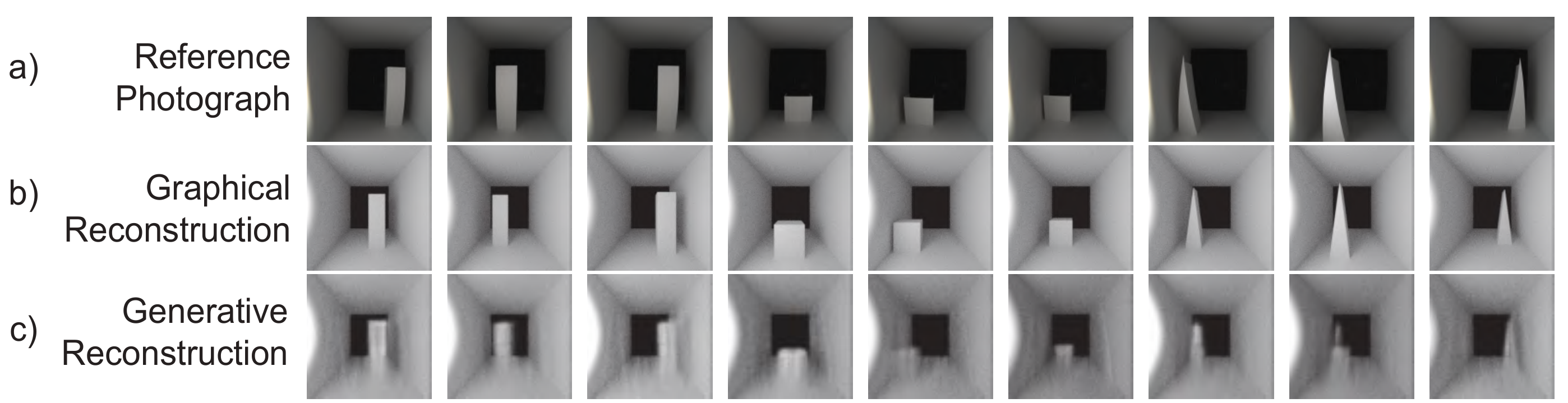}
\caption{Visualizing the occluded scene with a generative model. a) Photograph of the scene taken from the desired point-of-view, used for reference. b) Reconstruction results using a graphics program. The scene parameters are first estimated and then are used to render the image. c) Reconstruction results with a generative model. Different columns show results for different objects and positions from the test dataset.}
\label{fig:recons}
\end{figure*}

\subsubsection{Identification}
The test dataset was also used for identification accuracy. Fig.~\ref{fig:class}a shows the confusion matrix accounting for all locations. Fig.~\ref{fig:class}b shows the overall identification accuracy as a function of location (evaluated independently for each grid point). The identification accuracy is correlated with the size of the object (highest for the tall rectangular box and lowest for the short cube). This is expected as larger objects reflect more light to the floor and increase the measurement signal-to-noise ration. The identification accuracy across the entire grid is in the range of $[55.6,100]\%$ with a mean of 87.7\% (compared to 33.3\% random accuracy).

\section{Generative Occluded Scene Reconstruction}
\label{sec:genrative_model}

Our goal in this section is to go beyond the demonstrated computer vision tasks (localization, tracking, and identification) to full scene reconstruction. The recovered object location and class are sufficient to render an image of the hidden object using the graphical renderer. The success of this approach is limited by the accuracy of the localization and identification networks. Previous work by Klein et al.~\cite{Klein2016} have demonstrated a similar capability with a a consumer camera, but without object identification. Recent advances in generative models allow us to go beyond a graphics program and render a photo-realistic image of the hidden scene as if it was taken from a point of view not accessible to the camera (see Fig.~\ref{fig:overview}). 

The suggest approach is based on a variational auto-encoder (VAE)~\cite{kingma2013auto}. The VAE architecture is plotted in Fig.~\ref{fig:cnns}. The input to the generative model is the same as before (rectified, background subtracted image). The output is a rendering based on a point of view facing into the occluded area (Fig.~\ref{fig:overview}). 

Similar to the previous methods, the generative model can also be trained on rendered data. To that end, the desired output image is rendered along with the input image as part of the data generation pipeline described in section~\ref{sec:sim_datasets}. Thus, the training dataset is composed of sets of input and output renderings (75,000 data points). The generative model is trained for 100 epochs.

Figure~\ref{fig:recons} shows the results for the generative model evaluating its performance on the test dataset (section~\ref{sec:exp_res}). The figure compares: a) A reference photograph, taken with a camera at the desired position. b) A graphics rendering, using our data generation pipeline, with the estimated location and class as described in section~\ref{sec:exp_res}. c) The output of the generative model. The figure plots several examples with different object shapes and positions. The supplement provides more results and examples.

The main advantage of this generative approach is its generality since it can reconstruct the hidden scene without implicitly going through the localization and identification steps, and more importantly it does not require a labeled dataset, just pairs of images. This concept is further discussed and demonstrated in section~\ref{sec:lrg_gnrt}.

\section{Discussion}

\begin{figure}
\includegraphics[width=\linewidth]{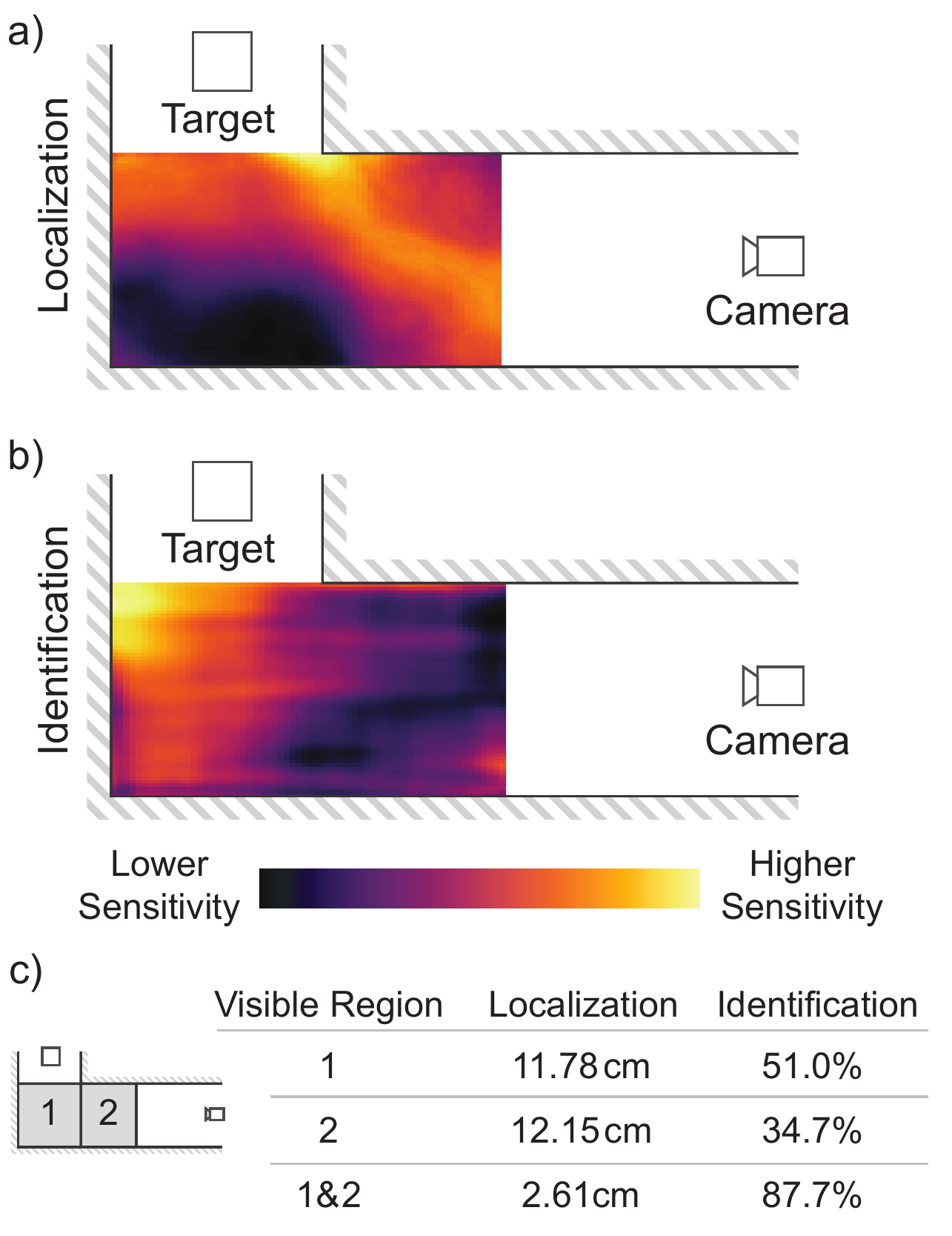}
\caption{Areas encoding information about the hidden scene. Brighter colors represent areas of higher sensitivity for the network in a) localization and b) identification. c) Comparing the importance of the area between the wall and corner (marked as 1 in the inset), and the area between the corner and the camera (marked as 2 in the inset). The table shows the localization and identification accuracies when only area 1 or 2 are available for the network. For localization both areas are equally important and using both significantly improves performance. For identification using just area 2 result in random prediction, and area 1 is essential for prediction.}
\label{fig:activations}
\end{figure}

\begin{figure}[t]
\includegraphics[width=\linewidth]{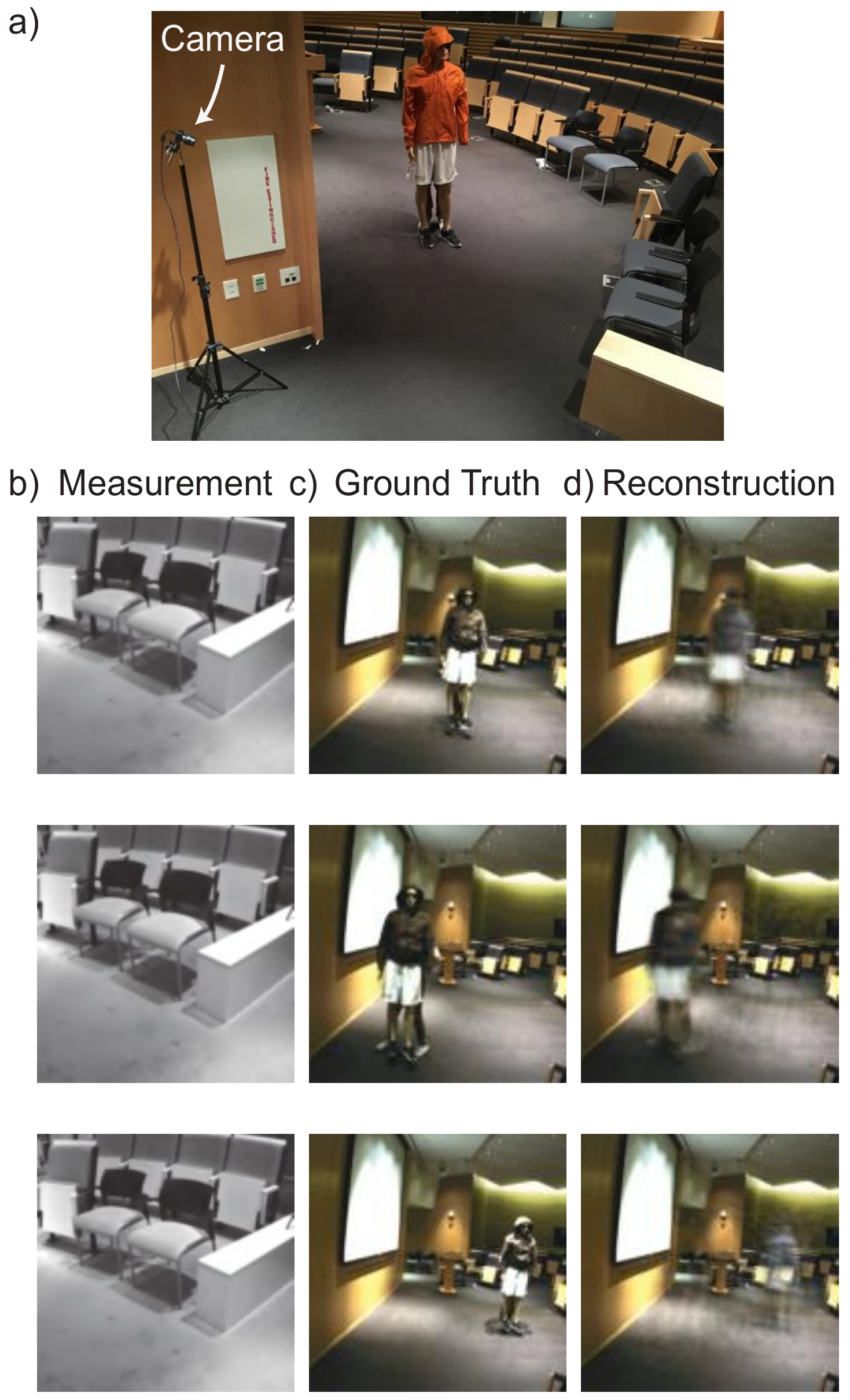}
\caption{Generative reconstruction results on trained real data. a) Scene photograph, showing the input camera (observing the floor) and the occluded scene. b) Input camera measurements. c) Ground truth measurements from the output camera (same camera used for training). d) Generative reconstruction, taking the background subtracted measurement as input and generate the scene from the desired point-of-view.}
\label{fig:real_recon}
\end{figure}

\subsection{Areas That Contain Information In The Scene}
It is important to consider which areas in the visible part of the scene encode information about the occluded part. Bouman et al.~\cite{Bouman_2017_ICCV} developed an analytical model showing that information about the angle of hidden objects are encoded in the area between the camera and the corner (marked as area 2 in Fig.~\ref{fig:activations}c). The input to our network includes a larger area that extends further, all the way to the wall facing the camera (marked as areas 1+2 in Fig.~\ref{fig:activations}c). 

To evaluate the contribution of different regions in the visible area for localization and identification accuracy, we block a small region ($20 \times 20$ pixels) in the input frame and try to localize and identify the hidden object (this is similar to ~\cite{zeiler2014visualizing}). Repeating this process for multiple objects, positions, and different blocked regions in the visible area provides a sensitivity map of the network (a total of 800 examples from the rendered dataset were used for this evaluation). Areas to which the network is more sensitive encode more information about the hidden object. 

The results for this sensitivity analysis for localization and identification is shown in Fig.~\ref{fig:activations}(a,b). For localization, substantial information is encoded around the occluding corner (as predicted by Bouman et al.~\cite{Bouman_2017_ICCV}). However, there is also significant sensitivity to the region that is between the corner to the further wall (area 1). For identification, most of the information is encoded beyond the corner in the area 1. To further evaluate this, we performed three tests: 1) only area 1 is visible, area 2 is blocked, 2) only area 2 is visible, area 1 is blocked (like in Bouman at al.), 3) both areas visible for baseline. We note that for localization when just one of the areas is visible the results are similar (equal contributions), however when both areas are available there is a dramatic boost in localization accuracy.  For identification the results are sharper; the use of area 2 alone results in random accuracy, while area 1 alone is able to identify with 51\% accuracy. It is expected that placing the separating line between the two areas in different locations would produce different results. We choose this particular separating line to compare our approach to Bouman et al. that used only area 2.

The model suggested by Bouman et al. requires a discontinuity in the the scene (such as a corner) in order to capture high frequency information that encodes the object's angle around the corner. Our approach is different, since we rely on a three-bounce light transport model. This model does not necessarily require such a discontinuity (although it helps), and can recover information about the hidden object from other regions in the visible scene. Furthermore, these regions contribute to our ability to localize the object in 2D and identify it. 

\subsection{Choosing Illumination Position and Camera Field-of-View}

In our setup, we choose to illuminate a wall facing the camera and observe the floor in front of the camera. This configuration is not unique. For example, if there is no wall facing the camera, it would be possible to illuminate the floor instead~\cite{Gariepy2015a}. In this case, it would be desirable to separate the camera field of view and the illuminated area to support the finite camera dynamic range. Other options may include illuminating the floor and observing the ceiling or vice-versa. 

\subsection{Generative Models Trained on Real Data}
\label{sec:lrg_gnrt}

The generative model demonstrated in section~\ref{sec:genrative_model} was trained on rendered data. Using rendered data is important in cases when it is hard to both gather and label datasets. However, the generative model approach does not require labels. Thus, it is possible to train such models on real data. The only requirement for such training data is the ability to access the occluded part of the scene to place a camera during training (we did not have such a requirement in our previous results). Such a requirement may be reasonable in the case of security cameras in places like alleys or museums. In such places it will be easy to gather a large dataset and create a system that provides security redundancy or reduced costs (less cameras).

To demonstrate this concept we placed two cameras in a conference room (see Fig.~\ref{fig:real_recon}). One camera is observing the floor, similar to the input camera in previous sections. The second camera is observing a scene outside the field of view of the first camera (Fig.~\ref{fig:real_recon}a). To train the system the cameras simply record synchronized videos that are then used to train the generative model. During the recordings a mannequin was moved around with different clothing in the occluded part of the room. A total of 14,600 frames were used for training (data gathering duration of less than 10 minutes). For testing we used a separate recording in which the mannequin was wearing different clothes. Several examples of the generative reconstruction are plotted in Fig.~\ref{fig:real_recon}.

\subsection{Main Limitations and Future Work}
Our work has two main limitations:

The first limitation of our approach is the limited number of predefined objects it can identify, and the ability to recover only a single object. In future work this limitation can be solved by training on a larger dataset with a more diverse set of objects. The CNNs developed here were trained on monochromatic images. Using color images may help to separate between different objects, and to ease object classification.

The second limitation of our approach is the requirement for known scene geometry. In future work this limitation can be solved by training on a larger dataset with a diverse set of scenes and geometries. An alternative is to scale the generative model option. By gathering real-world data from many locations, geometries, illumination conditions etc., it might be possible to train a generic generative model that is not constrained in a similar way. An alternative would be faster and more general physics based graphics renderers. Such systems would enable more complicated scenes and larger datasets.

\section{Conclusion}

Here, we present data-driven non-line-of-sight imaging with consumer flash photography. Different computational imaging tasks including object localization, tracking, identification and photo-realistic image reconstruction are demonstrated. While in the past, techniques with similar hardware have focused on geometric discontinuities, we showed that the visible scene encodes the occluded scene in other parts. In fact, limiting the analysis to one or two such discontinuities is sub-optimal. More complex scenes would encode more information about the hidden scene, and data-driven techniques are likely to play a key role in utilizing such information that may be very hard to model. This work is one example towards data-driven computational imaging.


{\small
\bibliographystyle{ieee}
\bibliography{egbib}
}

\newpage

\onecolumn

\huge \centerline{Supplementary Information}
\normalsize

\section*{1. Additional Results for Section 4 Generative Occluded Scene Reconstruction}

\begin{figure*}[h]
\includegraphics[width=\textwidth]{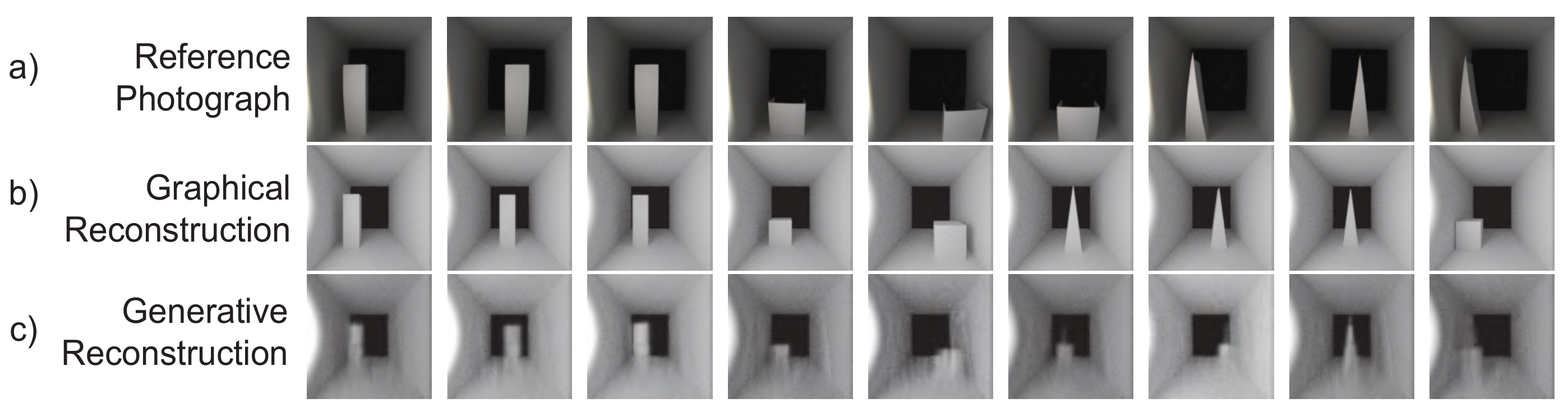}
\centering
\caption{Additional results to accompany Fig.~6 from the main text. Visualizing the occluded scene with a generative model. a) Photograph of the scene taken from the desired point-of-view, used for reference. b) Reconstruction results using a graphics program. The scene parameters are first estimated and then are used to render the image. c) Reconstruction results with a generative model. Different columns show results for different objects and positions from the test dataset.}
\label{fig:additional}
\end{figure*}

\section*{2. Implementation Details}
\begin{figure*}[h]
\includegraphics[width=\textwidth]{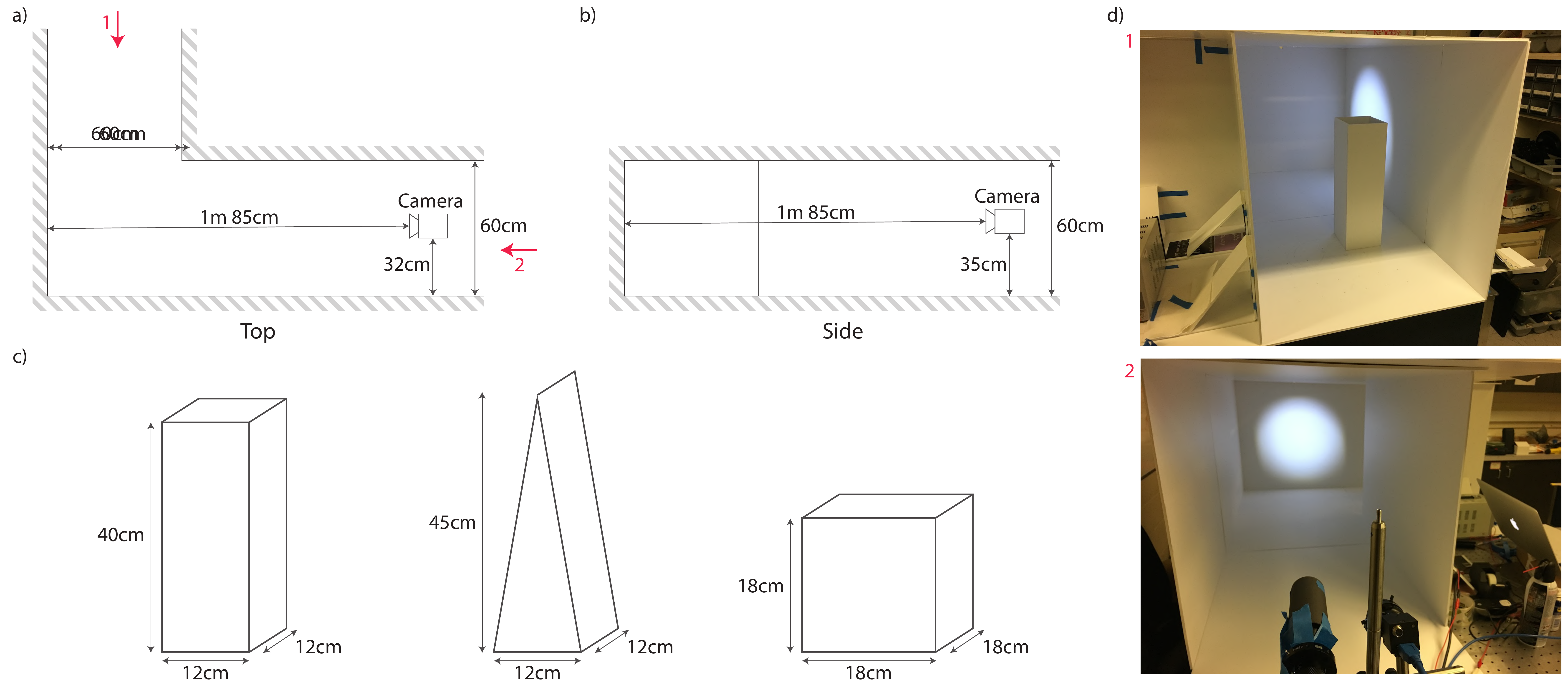}
\centering
\caption{Test scene. a) Dimensions of scene from an above perspective. b) Dimensions of scene from a side perspective. c) Dimensions of the three target objects tested. d) Photographs of the scene. Arrows in panel a point to the two perspectives.}
\label{fig:geometry}
\end{figure*}

The test scene and target objects were constructed out of poster board. The dimensions of the scene are shown in Fig.~\ref{fig:geometry}. A Flea3 FL3-U3-32S2C Point Grey camera was used to capture the data. The images were captured at 30fps with a 33ms exposure time. A LED flashlight was used to light the scene.

\section*{3. Rendering Details}
Blender 3d was automated to generate rendered training examples using the Cycles rendering engine. The same \textit{Principled BSDF} shader was used for the scene and target materials. The light was modeled as a spot source with multiple bounces. Model parameters were varied within a shifted uniform distribution: $X \left( \mu, \alpha \right) = \mu + Y$, where $Y \sim U \left[ -\alpha, \alpha \right]$ and are shown in Table \ref{table:parameters}.

\section*{4. CNN Details}
Table \ref{cnn} outlines the model architectures. $Conv2d(f,s,t)$ is a 2D convolution of size $s\times s$ with $f$ filters and a stride of $t$. $Pool$ is a $2\times2$ pooling layer. $FC(x,y)$ is a fully connected layer with an input size of $x$ and an output size of $y$. $BN$ is a batch normalization layer. $Drop$ is a dropout layer. $FC*$ is two fully connected layers that represent the mean an variance of the distribution. There is a ReLu after each $Conv2d$ and $FC$ layer.

The generative model trained on real data was a modified version of the generative model in the table. The input data was of size $128\times128$ instead of $64\times64$. The output data was $3\times128\times128$ instead of $64\times64$. To account for the dimension differences, the first and last $FC$ layers were enlarged. All other layers remained the same.

\begin{table}[]
\centering
\caption{Blender 3d renderer model parameters}
\label{table:parameters}
\begin{tabular}{lll}
\textbf{Scene Geometry} &  \\
- Ceiling Height & $G_h \sim X(61.5, 1)$ [cm] & \\
\textbf{Light (Spot Source)} &  & \\
- X Position & $L_x \sim X(185,3)$ [cm]\\
- Y Position & $L_y \sim X(28,3)$ [cm]\\
- Emission Strength & $L_e \sim X(2.6e^6,5e^5)$ \\
- Spot Shape Size & $L_{ss} \sim X(0.287,0.04)$ \\
- Spot Shape Blend & $L_{sb} \sim X(0.6,0.15)$ \\
\textbf{Camera} &  &\\
- Exposure & $C_e \sim X(1.1,0.2)$ & \\
\textbf{Materials (Principled BSDF)} & & \\
- Value & $M_v \sim X(0.8,0.2)$ & \\
- Specular &  $M_v \sim X(0.68,0.15)$ &\\
- Roughness & $M_v \sim X(0.5,0.2)$ & \\
\end{tabular}
\end{table}

\begin{table}[]
\centering
\caption{CNN Architectures}
\label{cnn}
\begin{tabular}{lll}
\textbf{Localization} & \textbf{Identification} & \textbf{Generative} \\
$Conv2d(8,5,1)$       & $Conv2d(8,5,1)$         & $Conv2d(8,5,1)$     \\
$Pool$                & $BN$                    & $Pool$              \\
$Conv2d(16,5,1)$      & $Pool$                  & $Conv2d(15,5,1)$    \\
$Pool$                & $Drop$                  & $Pool$              \\
$Conv2d(32,5,1)$      & $FC(8192,12)$           & $Conv2d(32,5,1)$    \\
$Pool$                & $Drop$                  & $Pool$              \\
$FC(2048,24)$         & $FC(12,3)$              & $FC(2048,100)$      \\
$FC(24,2)$            &                         & $FC*(100,20)$       \\
                      &                         & $FC(20,600)$        \\
                      &                         & $FC(600,4096)$     
\end{tabular}
\end{table}

\end{document}